\pdfoutput=1
\documentclass{llncs}
\usepackage{graphicx}
\usepackage{amsmath,graphicx,booktabs}
\usepackage{multirow}
\usepackage{epstopdf,url}
\usepackage{makeidx,amssymb}  
\usepackage{caption}
\usepackage{comment}
\usepackage{color}
\usepackage{algorithm}  
\usepackage{algpseudocode}
\usepackage[caption=false,font=footnotesize]{subfig}
\usepackage[dvipsnames]{xcolor}

\newcommand{\ve}[1]{\mathbf{#1}} 

\def\ie{\emph{i.e.}}
\def\eg{\emph{e.g.}}
\def\etal{{\em et al.}}

\begin{document}
\title{Uncertainty-aware Self-ensembling Model for Semi-supervised 3D Left Atrium Segmentation}

\author{Lequan Yu\inst{1}, Shujun Wang\inst{1}, Xiaomeng Li\inst{1},\\ Chi-Wing Fu\inst{1},  \and Pheng-Ann Heng\inst{1,2} }
\institute{Dept. of Computer Science and Engineering, The Chinese University of Hong Kong 
	\and T Stone Robotics Institute, The Chinese University of Hong Kong}
\maketitle            

\begin{abstract}
	Training deep convolutional neural networks usually requires a large amount of labeled data.
	However, it is expensive and time-consuming to annotate data for medical image segmentation tasks. 
	In this paper, we present a novel uncertainty-aware semi-supervised framework for left atrium segmentation from 3D MR images.
	Our framework can effectively leverage the unlabeled data by encouraging consistent predictions of the same input under different perturbations.
	Concretely, the framework consists of a student model and a teacher model, and the student model learns from the teacher model by minimizing a segmentation loss and a consistency loss with respect to the targets of the teacher model.
	We design a novel uncertainty-aware scheme to enable the student model to gradually learn from the meaningful and reliable targets by exploiting the uncertainty information.
	Experiments show that our method achieves high performance gains by incorporating the unlabeled data.
	Our method outperforms the state-of-the-art semi-supervised methods, demonstrating the potential of our framework for the challenging semi-supervised problems\footnote{Code is available in \url{https://github.com/yulequan/UA-MT}}.   
	
	\keywords{Semi-supervised learning \and Uncertainty estimation \and Self-ensembling \and Segmentation}
\end{abstract}

\section{Introduction}
Automated segmentation of left atrium (LA) in magnetic resonance (MR) images is of great importance in promoting the treatment of atrial fibrillation. 
With a large amount of labeled data, deep learning has greatly advanced the segmentation of LA~\cite{xiong2019fully}.
In the medical imaging domain, however, it is expensive and tedious to delineate reliable annotations from 3D medical images in a slice-by-slice manner by experienced experts.
Since unlabeled data is generally abundant, we focus on studying semi-supervised approach on LA segmentation by leveraging both limited labeled data and abundant unlabeled data.

Considerable effort has been devoted to utilizing unlabeled data to improve the segmentation performance in medical image community~\cite{bai2017semi,baur2017semi,chartsias2018factorised,ganaye2018semi,zhou2018semi}. 
For example, Bai~\etal~\cite{bai2017semi} introduced a self-training-based method for cardiac MR image segmentation, where the network parameters and the segmentation for unlabeled data were alternatively updated.
Besides, adversarial learning has been used in semi-supervised learning~\cite{dong2018unsupervised,nie2018asdnet,zhang2017deep}.
Zhang~\etal~\cite{zhang2017deep} designed a deep adversarial network to use the unannotated images by encouraging the segmentation of unannotated images to be similar to those of the annotated ones.
Another approach~\cite{nie2018asdnet} utilized an adversarial network to select the trustworthy regions of unlabeled data to train the segmentation network.
With the promising results achieved by self-ensembling methods~\cite{laine2016temporal,tarvainen2017mean} on semi-supervised natural image classification, Li~\etal~\cite{li2018semi} extended the $\Pi$-model~\cite{laine2016temporal} with transformation consistent for semi-supervised skin lesion segmentation.
Other approaches~\cite{cui2019semi,perone2018deep} utilized the weight-averaged consistency targets for semi-supervised MR segmentation.
Although promising progress has been achieved, these methods do not consider the reliability of the targets, which may lead to meaningless guidance.

In this paper, we present a novel uncertainty-aware semi-supervised learning framework for left atrium segmentation from 3D MR images by additionally leveraging the unlabeled data.
Our method encourages the segmentation predictions to be consistent under different perturbations for the same input, following the same spirit of mean teacher~\cite{tarvainen2017mean}.
Specifically, we build a teacher model and a student model, where the student model learns from the teacher model by minimizing the segmentation loss on the labeled data and the consistency loss with respect to the targets from the teacher model on all input data. 
Without ground truth provided in the unlabeled input, the predicted target from the teacher model may be unreliable and noisy.
In this regard, we design the uncertainty-aware mean teacher (UA-MT) framework, where the student model gradually learns from the meaningful and reliable targets by exploiting the uncertainty information of the teacher model.
Concretely, besides generating the target outputs, the teacher model also estimates the \textit{uncertainty} of each target prediction with Monte Carlo sampling. 
With the guidance of the estimated uncertainty, we filter out the unreliable predictions and preserve only the reliable ones (low uncertainty) when calculating the consistency loss.
Hence, the student model is optimized with more reliable supervision and in return, encourages the teacher model to generate higher-quality targets.
Our method was extensively evaluated on the dataset of MICCAI 2018 Atrial Segmentation Challenge. 
The results demonstrate that our semi-supervised method achieves large improvements for the LA segmentation by utilizing the unlabeled data, and also outperforms other state-of-the-art semi-supervised segmentation methods.

\section{Method}
\begin{figure*}[!t]
	\centering
	\includegraphics[width=1.0\linewidth]{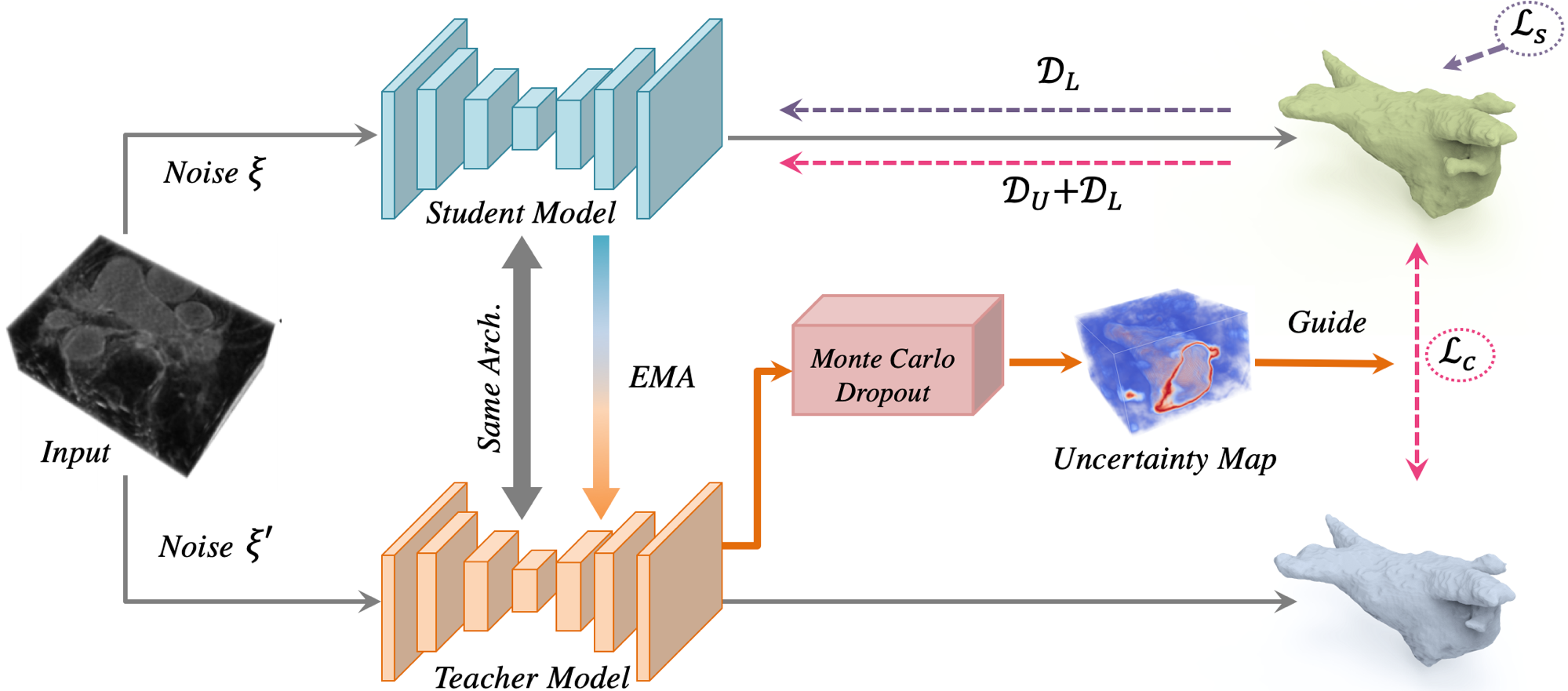}
	\caption{The pipeline of our uncertainty-aware framework for semi-supervised segmentation. The student model is optimized by minimizing the supervised loss $\mathcal{L}_s$ on labeled data $\mathcal{D}_L$ and the consistency loss $\mathcal{L}_c$ on both unlabeled data $\mathcal{D}_U$ and labeled data $\mathcal{D}_L$. 
		The estimated uncertainty from the teacher model guides the student to learn from the more reliable targets from the teacher.}
	\label{fig:pipeline}
	\centering
\end{figure*}

Fig.~\ref{fig:pipeline} illustrates our uncertainty-aware self-ensembling mean teacher framework (UA-MT) for semi-supervised LA segmentation.
The teacher model generates targets for the student model to learn from and also estimates the uncertainty of the target. 
The uncertainty-guided consistency loss improves the student model and the robustness of the framework.

\subsection{Semi-supervised Segmentation} 
We study the task of semi-supervised segmentation for 3D data, where the training set consists of $N$ labeled data and $M$ unlabeled data. 
We denote the labeled set as $\mathcal{D}_L = \left \{ (x_i, y_i) \right \}_{i=1}^{N}$ and the unlabeled set as~$\mathcal{D}_U = \left \{x_i \right \}_{i=N+1}^{N+M}$, where $x_i \in \mathbb{R}^{H \times W \times D}$ is the input volume and $y_i \in \{0,1\}^{H \times W\times D}$ is the ground-truth annotations. 
The goal of our semi-supervised segmentation framework is to minimize the following combined objective function:
\begin{equation}
\min_{\theta} \sum_{i=1}^{N} \mathcal{L}_s (f(x_i; \theta), y_i) + \lambda \sum_{i=1}^{N+M} \mathcal{L}_c (f(x_i; \theta', \xi'), f(x_i;\theta, \xi)),
\label{eq:semi}
\end{equation}
where $\mathcal{L}_s$ denotes the supervised loss (\eg, cross-entropy loss) to evaluate the quality of the network output on labeled inputs, and $\mathcal{L}_c$ represents the unsupervised consistency loss for measuring the consistency between the prediction of the teacher model and the student model for the same input $x_i$ under different perturbations. 
Here, $f(\cdot)$ denotes the segmentation neural network; $(\theta', \xi^{'})$ and $(\theta, \xi)$ represents the weights and different perturbation operations (\eg, adding noise to input and network dropout) of the teacher and student models, respectively. 
$\lambda$ is an ramp-up weighting coefficient that controls the trade-off between the supervised and unsupervised loss. 

Recent study~\cite{laine2016temporal,tarvainen2017mean} show that ensembling predictions of the network at different training process can improve the
quality of the predictions, and using them as the teacher predictions can improve the results.
Therefore, we update the teacher's weights $\theta'$ as an \textit{exponential moving average} (EMA) of the student's weights $\theta$ to ensemble the information in different training step~\cite{tarvainen2017mean}; see Fig.~\ref{fig:pipeline}. 
Specifically, we update the teacher's weights $\theta'_t$ at training step $t$ as: $\theta'_t = \alpha\theta'_{t-1} + (1-\alpha)\theta_t,$
where $\alpha$ is the EMA decay that controls the updating rate.

\subsection{Uncertainty-Aware Mean Teacher Framework}
Without the annotations in the unlabeled inputs, the predicted targets from the teacher model may be unreliable and noisy.
Therefore, we design an uncertainty-aware scheme to enable the student model to gradually learn from the more reliable targets.
Given a batch of training images, the teacher model not only generates the target predictions but also estimates the uncertainty for each target. 
Then the student model is optimized by the consistency loss, which focuses on only the confident targets under the guidance of the estimated uncertainty.

\subsubsection{Uncertainty Estimation.}
Motivated by the uncertainty estimation in Bayesian networks, we estimate the uncertainty with the Monte Carlo Dropout~\cite{kendall2017uncertainties}.
In detail, we perform $T$ stochastic forward passes on the teacher model under random dropout and input Gaussian noise for each input volume.
Therefore, for each voxel in the input, we obtain a set of softmax probability vector: $\{\ve{p}_t\}^T_{t=1}$.
We choose the predictive entropy as the metric to approximate the uncertainty, since it has a fixed range~\cite{kendall2017uncertainties}.
Formally, the predictive entropy can be summarized as:
\begin{equation}
\mu_c = \frac{1}{T}\sum_{t}\ve{p}^c_t \qquad \text{and} \qquad u =-\sum_{c}\mu_c \text{log} \mu_c,
\label{eq:uncertainty}
\end{equation}
where $\ve{p}^c_t$ is the probability of the $c$-th class in the $t$-th time prediction. Note that the uncertainty is estimated in voxel level and the uncertainty of the whole volume $U$ is $\{u\}\in \mathbb{R}^{H \times W \times D}$.

\subsubsection{Uncertainty-Aware Consistency Loss.}
With the guidance of the estimated uncertainty $U$, we filter out the relatively unreliable (high uncertainty) predictions and select only the certain predictions as targets for the student model to learn from. 
In particular, for our semi-supervised segmentation task, we design the uncertainty-aware consistency loss $\mathcal{L}_c$ as the voxel-level mean squared error (MSE) loss of the teacher and student models only for the most certainty predictions:
\begin{equation}
\mathcal{L}_c (f', f) = \frac{\sum_{v}\mathbb{I}(u_v<H)\left\| f'_v-f_v\right\|^2}{\sum_{v}\mathbb{I}(u_v<H)},
\label{eq:unsupervisedloss} 
\end{equation}
where $\mathbb{I}(\cdot)$ is the indicator function; $f'_v$ and $f_v$ are the predictions of teacher model and student model at the $v$-th voxel, respectively; $u_v$ is the estimated uncertainty $U$ at the $v$-th voxel; and $H$ is a threshold to select the most certain targets.
With our uncertainty-aware consistency loss in the training procedure, both the student and teacher can learn more reliable knowledge, which can then reduce the overall uncertainty of the model. 

\subsection{Technique Details}
We employ V-Net~\cite{milletari2016v} as our network backbone. We remove the short residual connection in each convolution block, and use a joint cross-entropy loss and dice loss~\cite{yang2017hybrid}.
To adapt the V-Net as a Bayesian network to estimate the uncertainty, two dropout layers with dropout rate 0.5 are added after the \textit{L-Stage 5} layer and \textit{R-Stage 1} layer of the V-Net.
We turn on the dropout in the network training and uncertainty estimation, while we turn off the dropout in the testing phase, as we do not need to estimate uncertainty. 
We empirically set the EMA decay $\alpha$ as $0.99$ referring to the previous work~\cite{tarvainen2017mean}.
Following~\cite{laine2016temporal,tarvainen2017mean}, we use a time-dependent Gaussian warming up function $\lambda(t)=0.1*e^{(-5(1-t/t_{max})^2)}$ to control the balance between the supervised loss and unsupervised consistency loss, where $t$ denotes the current training step and $t_{max}$ is the maximum training step.
Such design can ensure that at the beginning, the objective loss is dominated by the supervised loss term and avoid the network get stuck in a degenerate solution where no meaningful target prediction of unlabeled data is obtained~\cite{laine2016temporal}.
For the uncertainty estimation, we set $T=8$ to balance the uncertainty estimation quality and training efficiency.
We also use the same Gaussian ramp-up paradigm to ramp up the uncertainty threshold $H$ from $\frac{3}{4}U_{max}$ to $U_{max}$ in Eq.~\eqref{eq:unsupervisedloss}, where $U_{max}$ is the maximum uncertainty value (\ie, $\text{ln}2$ in our experiments). 
As the training continues, our method would filter out less and less data and enable the student to gradually learn from the relatively certain to uncertain cases.

\section{Experiments and Results}

\subsubsection{Dataset and Pre-processing.} We evaluated our method on the Atrial Segmentation Challenge dataset\footnote{http://atriaseg2018.cardiacatlas.org/}. 
It provides 100 3D gadolinium-enhanced MR imaging scans (GE-MRIs) and LA segmentation mask for training and validation. 
These scans have an isotropic resolution of $0.625\times 0.625 \times 0.625 \text{mm}^3$.
We split the 100 scans into 80 scans for training and 20 scans for evaluation. 
All the scans were cropped centering at the heart region for better comparison of the segmentation performance of different methods, and normalized as zero mean and unit variance. 

\subsubsection{Implementation.} 
The framework was implemented in PyTorch, using a TITAN Xp GPU. 
We used the SGD optimizer to update the network parameters (weight decay=$0.0001$, momentum=0.9).
The initial learning rate was set as 0.01 and divided by 10 every 2500 iterations. We totally trained 6000 iterations as the network has converged.
The batch size was 4, consisting of 2 annotated images and 2 unannotated images.
We randomly cropped $112\times112\times 80$ sub-volumes as the network input and the final segmentation results were obtained using a sliding window strategy.
We used the standard data augmentation techniques on-the-fly to avoid overfitting following~\cite{yu2017automatic}, including randomly flipping, and rotating with 90, 180 and 270 degrees along the axial plane.

\begin{table}[!t] 
	\centering
	\caption {Comparison between our method and various methods.} 
	\label{table:quanti_metric}
	\begin{tabular}{c|c|c|c|c|c|c}
		\toprule[2pt]
		\multirow{2}{*}{\bf{Method}} & \multicolumn{2}{c|}{\textbf{\# scans used}} & \multicolumn{4}{c}{\bf{Metrics}}\\
		\cline{2-7}					&Labeled 	&Unlabeled	&Dice[\%] 	 	&Jaccard[\%]		&ASD[voxel] 		&95HD[voxel] \\
		\hline
		Vanilla V-Net						&16		&0		&84.13			&73.26			&4.75			&17.93	\\
		Bayesian V-Net					 	&16		&0		&86.03			&76.06			&3.51			&14.26	\\
		\hline
		Vanilla V-Net 						&80		&0		&90.25			&82.40			&1.91			&8.29	\\
		Bayesian V-Net		 				&80		&0		&91.14			&83.82			&1.52			&5.75	\\
		\hline
		Self-training~\cite{bai2017semi}	&16		&64		&86.92			&77.28			&2.21			&9.19	\\
		DAN~\cite{zhang2017deep}			&16		&64		&87.52			&78.29			&2.42			&9.01	\\
		ASDNet~\cite{nie2018asdnet}			&16		&64		&87.90			&78.85			&\textbf{2.08}	&9.24	\\
		TCSE~\cite{li2018semi}				&16		&64		&88.15			&79.20			&2.44			&9.57	\\ \hline
		\textbf{UA-MT-UN (ours)}							&16		&64		&88.83			&80.13			&3.12			&10.04	\\	
		\textbf{UA-MT (ours)}				&16		&64		&\textbf{88.88}	&\textbf{80.21}	&2.26			&\textbf{7.32}\\
		\toprule[2pt]
	\end{tabular}
\end{table}

\subsubsection{Evaluation of Our Semi-supervised Segmentation.}
We use four metrics to quantitatively evaluate our method, including Dice, Jaccard, the average surface distance (ASD), and the 95\% Hausdorff Distance (95HD). 
Out of the 80 training scans, we use 20\% (\ie, 16) scans as labeled data and the remaining 64 scans as unlabeled data. 
Table~\ref{table:quanti_metric} presents the segmentation performance of V-Net trained with only the labeled data (the first two rows) and our semi-supervised method (UA-MT) on the testing dataset.
Compared with the Vanilla V-Net, adding dropout (Bayesian V-Net) improves the segmentation performance, and achieves an average Dice of 86.03\% and Jaccard of 76.06\% with only the labeled training data.
By utilizing the unlabeled data, our semi-supervised framework further improves the segmentation by 4.15\% Jaccrad and 2.85\% Dice.

To analyze the importance of consistency loss for labeled data and unlabeled data, we conducted another experiment (UA-MT-UN) with the consistency loss only on the unlabeled data.
The performance of this method is very close to UA-MT, validating that the performance of our method improves mainly due to the unlabeled data.
We trained the fully supervised V-Net with all 80 labeled scans, which can be regarded as the upper-line performance.
As we can see, our semi-supervised method is approaching the fully supervised ones.
To validate our network backbone design, we reference the state-of-the-art challenging method~\cite{chen2018multi}, which used multi-task U-Net for LA segmentation.
They reported a 90.10\% Dice on 20 testing scans with 80 training scans.
Compared with this method, we can regard our V-Net as a standard baseline model. 

\subsubsection{Comparison with Other Semi-supervised Methods.}
We implemented several state-of-the-art semi-supervised segmentation methods for comparison, including self-training based method~\cite{bai2017semi}, deep adversarial network (DAN)~\cite{zhang2017deep}, adversarial learning based semi-supervised method (ASDNet)~\cite{nie2018asdnet}, and $\Pi$-Model based method (TCSE)~\cite{li2018semi}.
Note that we used the same network backbone (Bayesian V-Net) in these methods for fair comparison.
As shown in Table~\ref{table:quanti_metric}, compared with the self-training method, the DAN and ASDNet improve by 0.60\% and 0.98\% Dice, respectively, showing the effect of adversarial learning in semi-supervised learning.
The ASDNet is better than DAN, since it selects the trustworthy region of unlabeled data for training the segmentation network.
The self-ensembling-based methods TCSE achieve slightly better performance than ASDNet, demonstrating that perturbation-based consistency loss is helpful for the semi-supervised segmentation problem.
Notably, our method (UA-MT) achieves the best performance over the state-of-the-art semi-supervised methods, except that the ASD performance is comparable with ASDNet, corroborating that our uncertainty-aware mean teacher framework has the full capability to draw out the rich information from the unlabeled data.

\begin{figure*}[!t]
	\centering
	\includegraphics[width=1.0\linewidth]{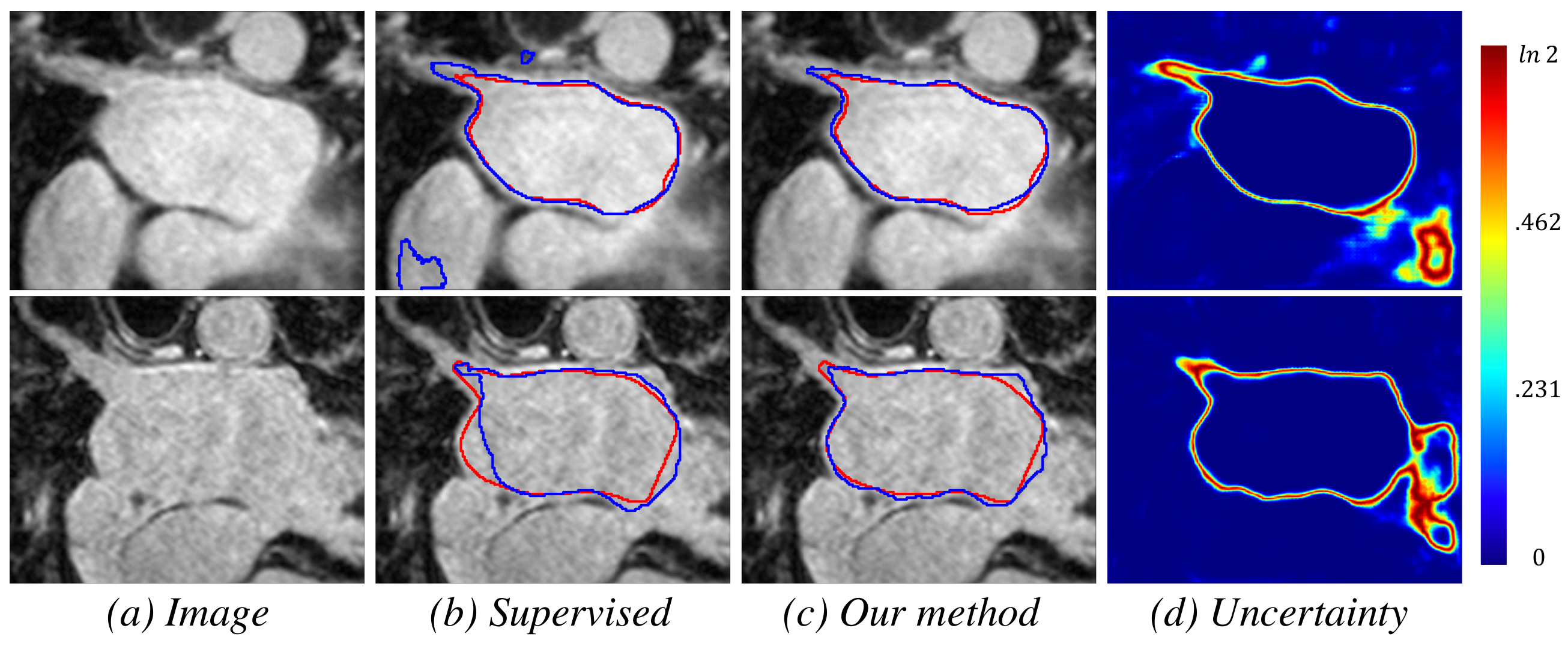}
	\caption{Visualization of the segmentations by different methods and the uncertainty. Blue and red colors show the predictions and ground truths, respectively.}
	\label{fig:seg_results}
	\centering
\end{figure*}

\begin{table}[!t] 
	\centering
	\caption {Quantitative analysis of our method.} 
	\label{table:differentdata}
	\begin{tabular}{c|c|c|c|c|c|c}
		\toprule[2pt]
		\multirow{2}{*}{\bf{Method}} & \multicolumn{2}{c|}{\textbf{\# scans used}} & \multicolumn{4}{c}{\bf{Metrics}}\\
		\cline{2-7}					&Labeled 	&Unlabeled	&Dice[\%] 	 	&Jaccard[\%]		&ASD[voxel] 		&95HD[voxel] \\
		\hline
		MT									&16		&64		&88.23			&79.29			&2.73			&10.64	\\	
		MT-Dice~\cite{cui2019semi}			&16		&64		&88.32			&79.37			&2.76			&10.50 	\\			
		Our UA-MT 								&16		&64		&88.88			&80.21			&2.26			&7.32	\\ \hline
		Bayesian V-Net					 	&8		&0		&79.99  		&68.12  		&5.48  			&21.11	\\
		Our UA-MT								&8		&72		&84.25  		&73.48  		&3.36  			&13.84	\\  \hline
		Bayesian V-Net					 	&24		&0		&88.52  		&79.70  		&2.60  			&10.45	\\
		Our UA-MT								&24		&56		&90.16  		&82.18  		&2.73  			&8.90	\\ 
		\toprule[2pt]
	\end{tabular}
\end{table}	

\subsubsection{Analysis of Our Method.}
To validate the effectiveness of our uncertainty-aware scheme, we evaluate the performance of the original mean teacher method (MT) and an adapted mean teacher method (MT-Dice) with dice-loss-like consistency loss~\cite{cui2019semi}.
As shown in Table~\ref{table:differentdata}, our uncertainty-aware method outperforms both the MT model and MT-Dice model. 
We also investigate the impact of using different numbers of labeled scans in our semi-supervised method.
As shown in Table~\ref{table:differentdata}, our semi-supervised method consistently improves the supervised-only V-Net (Bayesian V-Net) by utilizing the unlabeled data on both 10\% (\ie, 8) and 30\% (\ie, 24) labeled scans, demonstrating our method effectively utilizes the unlabeled data for the performance gains. 
In Fig.~\ref{fig:seg_results}, we show some segmentation examples of supervised method and our semi-supervised method, and the estimated uncertainty. 
Compared with the supervised method, our results have higher overlap ratio with the ground truth (the second row) and produce less false positives (the first row).
As shown in Fig.~\ref{fig:seg_results}(d), the network estimates high uncertainty near the boundary and ambiguous regions of great vessels.


\section{Conclusion}
We present a novel uncertainty-aware semi-supervised learning method for left atrium segmentation from 3D MR images.
Our method encourages the segmentation to be consistent for the same input under different perturbations to use the unlabeled data.
More importantly, we explore the model uncertainty to improve the quality of the target.
The comparison with other semi-supervised methods confirm the effectiveness of our method. 
The future works include investigating the effect of different uncertainty estimation manners and applying our framework to other semi-supervised medical image segmentation problems.
\\
\\
\textbf{Acknowledgments.}
The work was partially supported by HK RGC TRS project T42-409/18-R and
in part by the CUHK T Stone Robotics Institute, The Chinese University of Hong Kong.

\bibliographystyle{splncs04}
\bibliography{paper182}

\end{document}